# Persona and Persuasive Framing in AI Voice Agents: A 2×2 Field Experiment with Children

*Thilo Tamme*[*]
Technical University of Munich
thilo.tamme@tum.de

*David Steck*[*]
Technical University of Munich
david.steck@tum.de

*Anton Hantel*
Massachusetts Institute of Technology
hantel@mit.edu

## Abstract

*Conversational agents increasingly interact with children, yet evidence on how their design shapes children's susceptibility to persuasion comes almost entirely from the lab. We report a 2×2 randomized field experiment embedded in a public German Santa Claus telephone hotline. Children's calls were randomly routed to one of four LLM voice agents varying persona (Santa, high authority, vs. Helper, low authority) and framing (persuasive nudges toward prosocial wishes vs. neutral). Of 1,072 logged calls, 89 conversations (median age 6) met inclusion criteria. Persuasive framing raised the probability of a prosocial wish from 11.6% to 45.7%, robust to controls. Persona authority showed a near-zero effect: Santa did not outperform the Helper. Persona instead shaped engagement; children hung up on the Helper far more often within the first minute (65% vs. 39%). Where context already lends an agent legitimacy, how it speaks shapes children's compliance more than who it claims to be.*

**Keywords:** conversational agents, children, persuasion, field experiment, agent persona

## 1. Introduction

Voice-based conversational agents (CAs) have moved from novelty to normal in children's lives. Children talk to smart speakers at home, attribute intelligence and personality to them, and form parasocial bonds with their voices (Druga et al., 2017; Hoffman et al., 2021). At the same time, large language models (LLMs) have made CAs vastly more fluent and, as a rapidly growing literature shows, measurably persuasive with adults (Lin et al., 2025; Hackenburg et al., 2025). Combined with evidence that children tend to overtrust AI systems (Solyst et al., 2024) while their persuasion knowledge is still developing (Friestad & Wright, 1994; Watson et al., 2024), CAs ability to persuade adults suggests that design choices modulating a CA's persuasive influence on children deserve empirical scrutiny; regulators have begun to act on this concern (European Union, 2024, Art. 5; Federal Trade Commission, 2025).

Two design levers are especially salient for human-like agents. The first is *persona*: who the agent claims to be (Santa Claus vs. Santa's Helper) (Pornpitakpan, 2004). The second is *framing*: what the agent is scripted to pursue in conversation. Recent large-scale work suggests that prompt-level framing and message strategy, rather than model scale or personalization, drive most of LLM persuasion (Hackenburg et al., 2025).

We test both levers in the field with young children. We embedded a 2×2 randomized experiment in a public Santa hotline operated before Christmas 2025. Calling children were randomly routed to an LLM voice agent that introduced itself either as Santa Claus; an archetypal, culturally legitimate authority figure (Goldstein & Woolley, 2016); or as Santa's Helper; both either conversed neutrally about Christmas wishes or gently nudged the child toward wishes that benefit someone else (prosocial wishes). The outcome variable is whether the child spontaneously expressed a prosocial wish during the call.

The study makes three contributions. First, it provides rare in-the-wild causal evidence on child–agent persuasion: framing increased prosocial wish expression roughly four-fold, while persona authority showed no detectable effect on what children wished, but a pronounced effect on whether callers stayed in the conversation at all. Second, the study contributes methodologically: a replicable, ethically constrained field paradigm built on consent-gated access, a fictional rather than institutional authority figure, a cap on nudges, and anonymized data, paired with a transcript-based manipulation check that many field studies omit. Third, it quantifies the screening funnel of public child-directed deployments: fewer than one in ten logged calls yielded an analyzable child conversation. We report this yield, call duration, and child speech share as planning parameters for future field studies.

[*] The authors have contributed equally.

## 2. Theoretical background

### 2.1. Agents as social actors

The Computers-Are-Social-Actors (CASA) paradigm holds that people mindlessly apply social rules and categories (politeness, gender, authority) to machines that emit even minimal social cues (Nass et al., 1994; Nass & Moon, 2000; Reeves & Nass, 1996). Voice is among the strongest of these cues, carrying identity, warmth, and authority information (Seaborn et al., 2021), and contemporary extensions of CASA argue that richer media scripts now shape these responses (Gambino et al., 2020). Anthropomorphic design reliably increases trust and compliance intentions in service contexts (Blut et al., 2021), and identity claims by voice agents causally affect behavior in the field (Luo et al., 2019; Xu et al., 2024).

Children are a boundary-testing population for CASA. They attribute intelligence and feelings to voice assistants (Druga et al., 2017; Andries & Robertson, 2023), hold ontologically ambivalent views of what such agents are (Xu & Warschauer, 2020), calibrate trust in them in age-graded ways (Girouard-Hallam & Danovitch, 2022), and form parasocial relationships with their characters (Hoffman et al., 2021).

### 2.2. Persona authority and source effects

Authority is a canonical persuasion principle (Cialdini, 2021; Milgram, 1963), and source credibility research documents that identical messages persuade more from credible sources (Hovland & Weiss, 1951; Pornpitakpan, 2004). For embodied agents, however, authority is double-edged: social robots displaying formal authority can trigger reactance and persuade *less* when their authority lacks legitimacy (Saunderson & Nejat, 2021). Santa Claus is a theoretically interesting persona because his authority is culturally legitimate and benevolent for believing children: belief peaks around ages 4–8 (Goldstein & Woolley, 2016) and declines around age 8 (Mills et al., 2024). We therefore expected the Santa persona to elicit more prosocial wishes than a low-authority assistant persona under persuasive framing (H2).

### 2.3. Framing, peripheral route, and age

The Elaboration Likelihood Model (ELM) distinguishes effortful central-route processing from peripheral-route processing driven by cues and heuristics (Petty & Cacioppo, 1986); chatbot studies confirm both routes operate in CA persuasion (Chen et al., 2025). Young children interacting with a legitimately perceived agent in a playful context are unlikely to centrally elaborate, allowing subconscious cues, such as trust to flow through the peripheral route. Consistent with this, recent evidence suggests that *how* an LLM agent is prompted to converse outweighs *who* it claims to be or how large the model is (Hackenburg et al., 2025), and how chatbot empathy and identity interact in donation contexts (Park et al., 2023). Laboratory work shows social robots can increase children's prosocial behavior (Peter et al., 2021). We therefore expected the persuasive framing to increase children's prosocial wish expression relative to neutral conversation (H1). Because persuasion knowledge, the ability to recognize and discount influence attempts, develops with age (Friestad & Wright, 1994; John, 1999; Watson et al., 2024), we expected younger children to be more susceptible to the persuasive framing (H3).

**H1.** Children in the persuasive framing condition are more likely to express a prosocial wish than children in the neutral condition.

**H2.** The high-authority Santa persona elicits more prosocial wishes than the low-authority Helper persona under persuasive framing.

**H3.** Younger children are more susceptible to the persuasive framing than older children.

Field evidence on these questions is scarce. HRI scholars have long argued that the field's theories rest too heavily on laboratory studies (Jung & Hinds, 2018), and the most informative recent CA experiments with behavioral outcomes are field studies with adults (Luo et al., 2019; Xu et al., 2024). To our knowledge, no prior field experiment has manipulated persona authority and persuasive framing with children and a behavioral outcome.

## 3. Methodology

### 3.1. Setting and design

The study was embedded in a public German-language Santa hotline operated by a seasonal service provider *WeihnachtsmannWerk* and promoted through local press, radio, and national online media. The hotline served a predominantly adult user base: parents obtained the number through the operator's website and email communication, and access required an adult to consent on the website to research data collection. Reaching the line therefore required adult initiation, not independent dialing by children. Each incoming call was randomly forwarded by a telephony routing function to one of four LLM voice agents (built on a commercial voice-agent platform, LLM: Gemini 2.5 Flash Lite), implementing a 2×2 between-subjects design: persona (Santa Claus vs. Santa's Helper) × framing (persuasive vs. neutral). All agents shared identical guardrails, age-appropriate conversational goals, and the same technical stack; only the persona introduction and the framing script differed. In the persuasive condition, agents were prompted to deliver at most two gentle nudges inviting a wish that benefits someone else (e.g., "Is there something you would wish for someone else?"). After each call, consenting callers received a parental survey link.

Ethical constraints were designed into the deployment: consent gating, a fictional rather than institutional authority figure, a hard cap on nudge frequency, age-appropriate guardrails, the child's ability to end the interaction by hanging up at any moment, and anonymized conversation identifiers. The design follows child-centered AI guidance (UNICEF, 2021) and avoids the manipulative techniques prohibited by Article 5 of the EU AI Act (European Union, 2024).

## 3.2. Sample

Between November 2, 2025 and January 21, 2026, the hotline logged 1,072 calls. Exclusions were applied sequentially (Figure 1): 578 calls fell outside the December 15–25 study window (the pre-window period served setup and testing with a single Santa agent outside of the 2x2 design; post-Christmas calls contained no meaningful wish interactions); 259 lasted one minute or less and could not have reached the treatment stage; 123 were repeat calls from the same number (only the first call was retained); 18 callers were adults speaking on their own behalf (access was adult-mediated, so some calls came from parents rather than children); four calls were unintelligible due to technical failure; one was a jailbreak attempt. The analytical sample comprises 89 conversations (8.3% of logged calls)Within the study window, calls were routed to the four agents in balanced proportions (Table 1; $\chi^2(3) = 1.30$, $p = .73$), and exclusion was unrelated to framing ($\chi^2(1) = 0.01$, $p = .92$). This is expected by design: nudges were delivered only after the first minute, so early hang-ups cannot reflect the framing manipulation. Exclusion did, however, differ by persona; Section 4.5 examines this pattern. The final cells (Table 2) were Santa-neutral (n = 28), Santa-persuasive (n = 26), Helper-neutral (n = 15), Helper-persuasive (n = 20). A usable stated age was available for 70 children (median 6, range 3–15); 56 children were nine or younger. Calls lasted 1.0–10.0 minutes (median 2.2).

**Table 1. Call attrition by condition within the study window (December 15–25).**

| Condition | Calls | ≤1m | Dup. | Oth. | Incl. |
|---|---|---|---|---|---|
| Santa neutral | 118 | 43 | 40 | 7 | 28 |
| Santa persuasive | 119 | 50 | 38 | 5 | 26 |
| Helper neutral | 123 | 79 | 24 | 5 | 15 |
| Helper persuasive | 134 | 87 | 21 | 6 | 20 |
| **Total** | **494** | **259** | **123** | **23** | **89** |

**Table 2. Experimental cells, automated nudge detection, and prosocial wish expression (N = 89).**

| Condition | n | ≥1 nudge | M nudges (SD) | Prosocial wish |
|---|---|---|---|---|
| Santa neutral | 28 | 0 (0%) | 0.00 (0.00) | 4 (14.3%) |
| Helpe neutral | 15 | 1 (6.7%) | 0.07 (0.26) | 1 (6.7%) |
| Santa persuasive | 26 | 21 (80.8%) | 1.23 (0.99) | 12 (46.2%) |
| Helper persuasive | 20 | 14 (70.0%) | 1.25 (1.12) | 9 (45.0%) |
| **Total** | **89** | **36 (40.4%)** | **0.65 (0.97)** | **26 (29.2%)** |

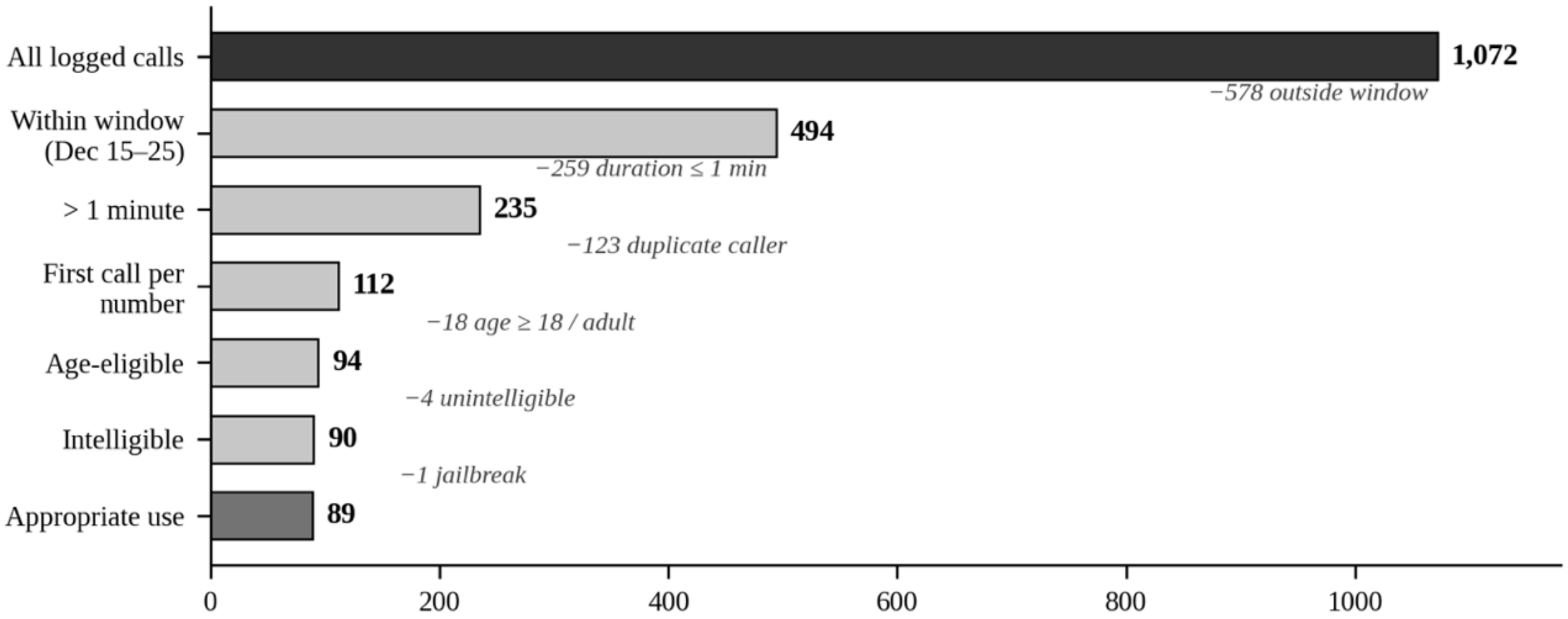

**Figure 1. Sequential exclusion of the 1,072 logged calls down to the analytical sample of 89 conversations.**

### 3.3. Measures and coding

The dependent variable is binary: whether the child expressed a prosocial wish, defined as a wish whose primary beneficiary is someone other than the child (e.g., health for a parent, peace, the family being together; cf. Peter et al., 2021). Transcripts were coded manually; a second, independent reconciliation pass coded conversations added during data cleaning, with each conversation's coding source documented. As a robustness and fidelity check, we additionally ran a lexical analysis over all transcripts that (a) detects prosocial nudges in agent turns and (b) conservatively flags other-oriented wish expressions in child turns. Agreement between the manual coding and the conservative automated classifier was 87% (Cohen's κ = .73), with the few disagreements concentrated in borderline family-togetherness wishes.

### 3.4. Analysis

Given the binary outcome, hypotheses were tested with logistic regressions (Table 3): H1 via the persuasive-condition main effect (full sample and core sample of children aged nine or younger), with a robustness model adding nudge count, age, call duration, and days remaining until Christmas; H2 via cell-level contrasts and a direct contrast among the persuasive agents; H3 via the age effect. We report Wald z statistics; Fisher's exact tests corroborated all binary contrasts.

**Table 3. Logistic regression results (Wald tests; manual outcome coding).**

| Model | Term | β (SE) | p |
|---|---|---|---|
| H1 full (N = 89) | Persuasive | 1.85 (0.56) | .001 |
| H1 core ≤ 9 (n = 56) | Persuasive | 2.35 (0.73) | .001 |
| H1 robust (n = 70) | Persuasive | 2.74 (0.89) | .002 |
| | Nudges | 0.03 (0.42) | .95 |
| | Age | −0.11 (0.12) | .36 |
| | Duration (min) | 0.33 (0.21) | .12 |
| | Days to Christmas | 0.34 (0.16) | .037 |
| Cells (ref. Santa–neutral) | Helper neutral | −0.85 (1.17) | .47 |
| | Santa persuasive | 1.64 (0.67) | .014 |
| | Helper persuasive | 1.59 (0.70) | .024 |
| H2 contrast (persuasive only) | Santa persona | 0.05 (0.60) | .94 |
| H3 (n = 70) | Age | 0.06 (0.09) | .47 |

## 4. Results

### 4.1. Manipulation check

Automated nudge detection verified treatment delivery: at least one prosocial nudge was detected in 76% of persuasive-condition conversations (35/46; M = 1.24 per call) versus one borderline instance in 43 neutral conversations (Mann–Whitney U test, $p < .001$). Persuasive calls without a detected nudge were predominantly conversations that ended before the script reached the nudging stage. Contamination of the control condition was negligible.

### 4.2. Persuasive framing (H1)

Children in the persuasive condition expressed prosocial wishes in 45.7% of conversations (21/46), versus 11.6% (5/43) under neutral framing (Figure 2), an odds ratio of 6.4 (Fisher's exact $p < .001$). The logistic regression confirmed the effect in the full sample (β = 1.85, SE = 0.56, z = 3.31, p = .001; 95% CI for the odds ratio [2.1, 19.1]) and in the core sample (β = 2.35, SE = 0.73, p = .001). The effect was robust to controls for nudge count, age, duration, and days to Christmas (β = 2.74, SE = 0.89, p = .002); among the controls, only days remaining until Christmas reached significance (β = 0.34, p = .037), with earlier-window calls more likely to contain prosocial wishes. Within the persuasive condition, conversations in which at least one nudge was actually delivered showed prosocial wishes in 54% of cases against 18% when no nudge was delivered, although the trend across nudge counts was not significant (Spearman ρ = .22, p = .15). H1 is supported.

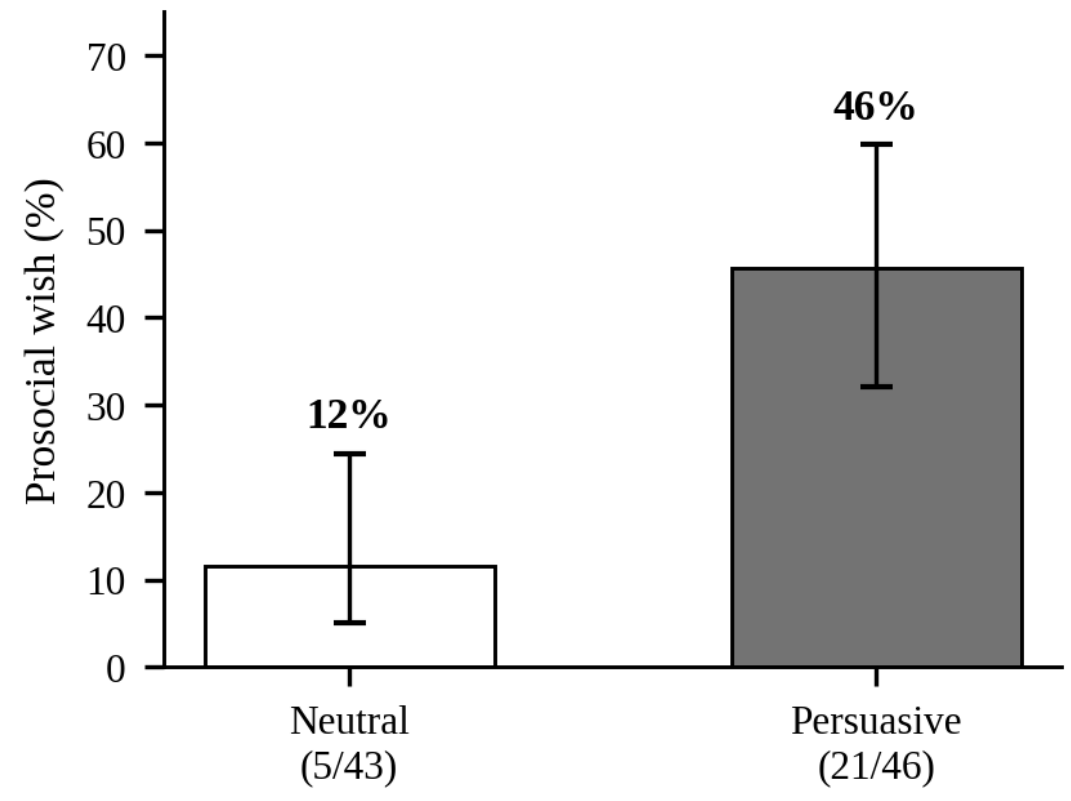

**Figure 2. Prosocial wish expression by framing (95% Wilson CIs).**

## 4.3. Agent persona (H2)

Both persuasive cells outperformed the Santa-neutral baseline (Santa-persuasive: 46.2%, β = 1.64, p = .014; Helper-persuasive: 45.0%, β = 1.59, p = .024; Table 3), but the two persuasive personas were statistically indistinguishable, with a near-zero point estimate (β = 0.05, SE = 0.60, p = .94). Descriptively, the low-authority Helper matched the high-authority Santa almost exactly (Figure 3). Among included calls, Helper conversations even ran descriptively longer (median 2.5 vs. 1.9 minutes; Mann–Whitney U test, p = .063), though included Helper calls are a more strongly selected subsample (Section 4.5). H2 is not supported. Persona did, however, strongly affect whether callers engaged with the agent at all (Section 4.5).

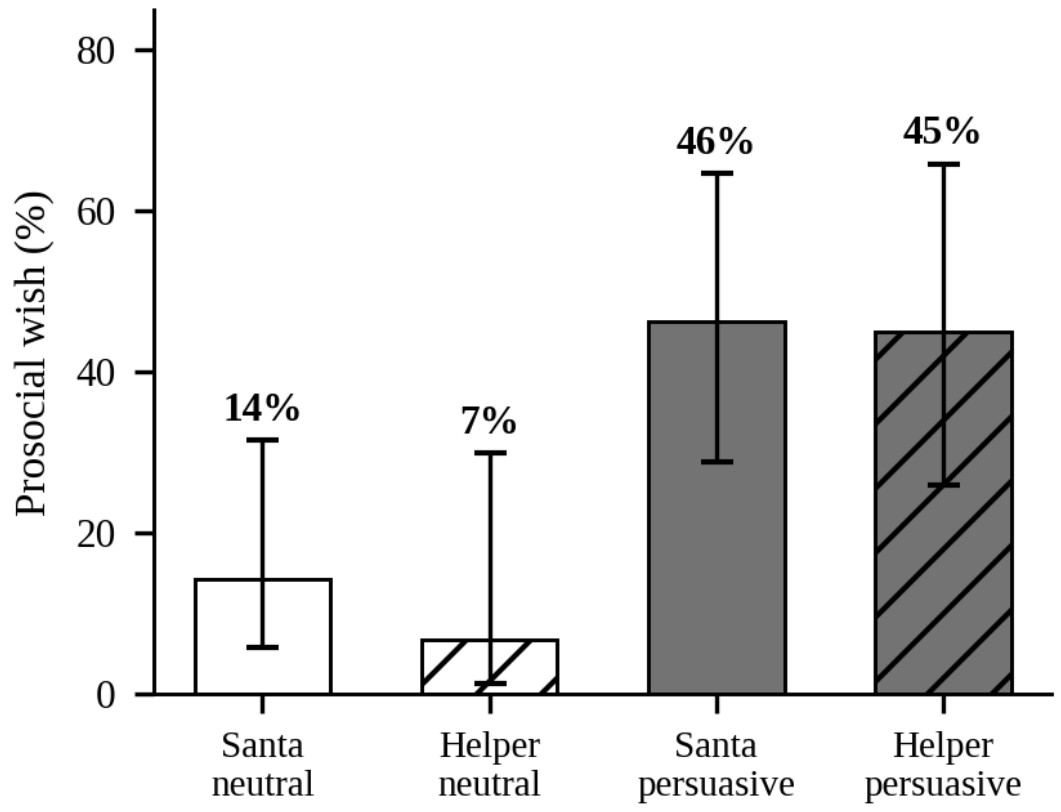

**Figure 3. Prosocial wish expression by experimental cell (95% Wilson CIs).**

## 4.4. Age (H3)

Age did not predict prosocial wish expression (β = 0.06, SE = 0.09, p = .47), nor did it reach significance in any multivariate model. H3 is not supported. A post-hoc simulation indicates the persona and age contrasts were underpowered: detecting the observed persona difference with 80% power would require roughly 160 observations per group. The null findings are therefore indicative.

## 4.5. Persona and early disengagement

The exclusion step removing calls of one minute or less was strongly persona-dependent. Within the study window, 64.6% of Helper calls ended within the first minute, versus 39.2% of Santa calls (OR = 2.82, 95% CI [1.96, 4.07], $\chi^2(1) = 31.8$, p < .001). The gap was equally present under both framings (interaction p = .59) and held across alternative cutoffs from 30 to 120 seconds. It was driven by immediate disengagement: 40.5% of Helper callers hung up without saying an intelligible word, against 8.9% of Santa callers (OR = 6.99, 95% CI [4.19, 11.67], p < .001); all of these calls were ended by the caller. Figure 4 shows the resulting pattern: the personas diverge during the greeting and converge again beyond two minutes: the persona difference is confined to the opening of the call. Because nudges had not yet been delivered at this stage, the analysis is unconfounded by the framing manipulation, though it remains exploratory.

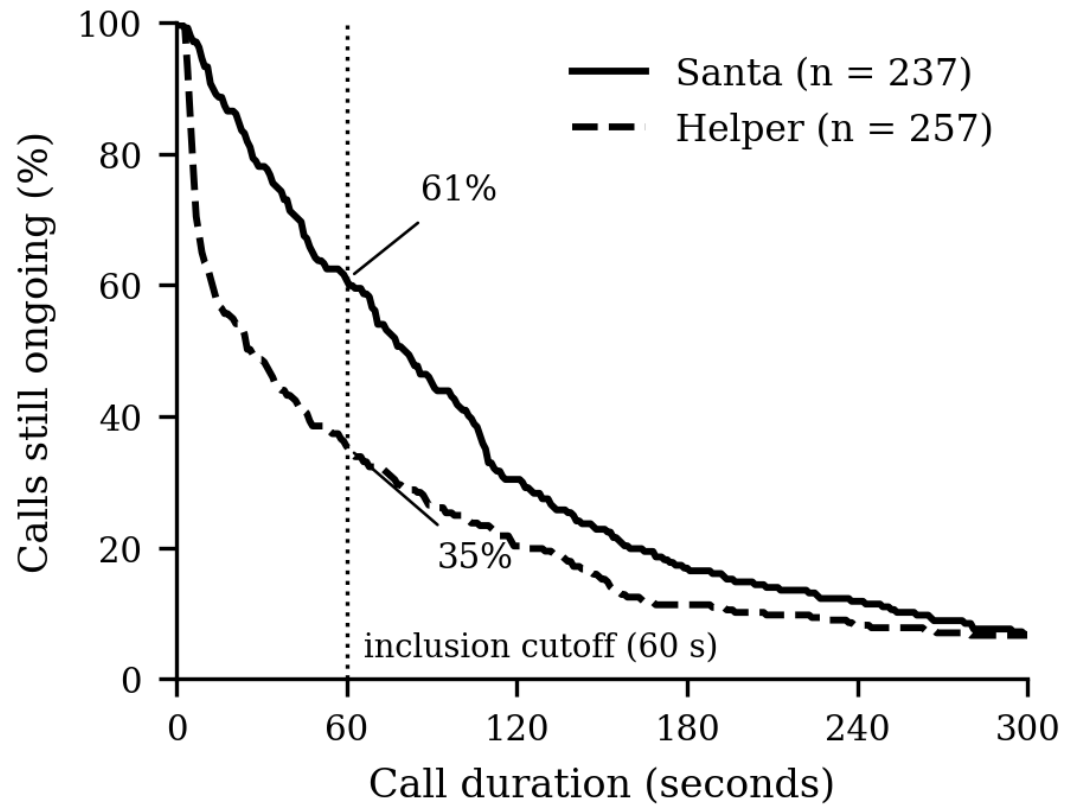

**Figure 4. Share of calls still ongoing by duration and persona (all study-window calls, n = 494). Nudges were delivered only after the first minute.**

# 5. Discussion

## 5.1. Framing as the primary lever

The pattern of a large, robust framing effect alongside a near-zero persona point estimate is informative even under limited power. Under the ELM, young children in a ritualized, emotionally charged context process peripherally (Petty & Cacioppo, 1986; Chen et al., 2025): the situation itself (calling Santa's hotline at Christmas) supplies legitimacy, and the agent's conversational moves do the persuading. Once a child is inside this legitimate frame, we found no evidence that *who* delivers the nudge adds persuasive force: a famous authority figure and his unnamed assistant achieved practically identical rates. This matches adult evidence that prompt-level strategy outweighs persona and personalization in LLM persuasion (Hackenburg et al., 2025) and with the finding that authority cues can backfire when their legitimacy is questionable (Saunderson & Nejat, 2021). For the digital-humans literature, this suggests a boundary condition for celebrity-persona effects (Yuan et al., 2023): when the context already carries legitimacy, persona authority may add little.

The attrition pattern (Section 4.5) adds a second stage to this account. Persona did not change compliance among children who engaged, but it strongly shaped who engaged: 65% of Helper calls ended within the first minute against 39% of Santa calls, and 41% versus 9% ended without the caller speaking at all. Authority did matter by keeping callers in the conversation (but not by amplifying the nudge). Under CASA, ending a conversation with a high-authority figure may carry social costs that ending one with an anonymous assistant does not (Nass & Moon, 2000; Cialdini, 2021). Two rival explanations remain. Callers had dialed a Santa hotline and may have disengaged when an assistant was not who they expected, and the personas differed in voice as well as identity. We cannot separate deference to authority from either. Either way, persuasion here is two-staged: persona governed who engaged, framing governed compliance once they did.

A measurement caveat bears on the null. For these children the Helper may not have been low-authority at all, but an emissary of the same institution, inheriting Santa's authority by association; high belief in Santa (M = 4.16/5) and low AI awareness (M = 2.21/5) make this plausible. If so, we shifted persona but not authority, and the equal persuasion reflects a weak contrast, not a real absence of an authority effect. Taken with the engagement gap, this points back to the section's claim: legitimacy sat in the frame, not the figure, with persona doing its work at the threshold. Separating persona authority from institutional legitimacy is the clear next step.

## 5.2. Implications for design and policy

Practically, the results imply that the persuasive power of child-directed CAs resides primarily in their scripting, which is cheap to change, easy to A/B test, and invisible to parents, rather than in their visible persona. The asymmetry matters for governance: persona is auditable at a glance; framing is not. Our automated nudge detection demonstrates that conversation-level auditing is feasible: simple lexical classifiers recovered the treatment script reliably and achieved acceptable agreement with manual outcome coding (κ = .73).

The EU AI Act prohibits manipulative AI techniques that exploit age-based vulnerability (European Union, 2024, Art. 5), and the FTC has opened an inquiry into companion chatbots and minors (Federal Trade Commission, 2025). Roughly 80% of apps used by young children already deploy manipulative design (Radesky et al., 2022), and LLMs simulate empathy without understanding (Kurian, 2024). We chose a deliberately benign outcome, prosocial wishes, as an ethically defensible stand-in for persuasion that could just as easily serve commercial ends (Susser et al., 2019). Catching the hidden version likely requires exactly this kind of transcript-level scrutiny.

## 5.3. Limitations and future research

Four limitations bound our claims. First, despite more than 1,000 logged calls, the analytical sample is small (n = 89) and underpowered for persona and age contrasts; the persona contrast additionally rests on a sample selected by persona-dependent early hang-ups (Section 4.5). Second, stated ages are unverified, and parents were present in most calls, which may dampen or amplify treatment effects. Third, the single-culture, single-ritual setting and the masked AI identity limit generalization to disclosed-AI settings, where disclosure reduces persuasion (Luo et al., 2019; Xu et al., 2024). Fourth, persuasion was measured within a single call; durable attitude or behavior change is unknown.

# 6. Conclusion

In a randomized field experiment inside a real Santa hotline, persuasive conversational framing roughly quadrupled the share of children who expressed a prosocial wish. Persona worked differently. Among children who stayed on the line, we found no evidence that a high-authority figure was more persuasive than a low-authority assistant: the script, not the speaker, moved the wish. But persona shaped who stayed at all, with far more callers hanging up on the Helper before any nudge landed. Persuasion ran in two stages: persona chose who listened, framing chose what they wished for.

The asymmetry matters beyond Christmas. The persuasive power of a child-directed agent lived in its conversation design, the layer cheapest to change and hardest for a parent or regulator to see. That layer is at least open to scrutiny: a simple classifier could reliably tell which script a child had heard. We studied a benign wish, but the mechanism is indifferent to the wish it serves. As conversational agents become part of childhood, both the promise and the risk live there, and bringing them into view is where this work begins.